\documentclass[sn-mathphys,Numbered]{sn-jnl}

\usepackage{graphicx}%
\usepackage{multirow}%
\usepackage{amsmath,amssymb,amsfonts}%
\usepackage{amsthm}%
\usepackage{mathrsfs}%
\usepackage[title]{appendix}%
\usepackage{xcolor}%
\usepackage{textcomp}%
\usepackage{manyfoot}%
\usepackage{booktabs}%
\usepackage{algorithm}%
\usepackage{algorithmicx}%
\usepackage{algpseudocode}%
\usepackage{listings}%
\usepackage{tabularx}
\usepackage{lineno}

\begin{document}
\title[\textcolor{black}{When Large Language Models Meet Evolutionary Algorithms: Potential Enhancements and Challenges}]{\textcolor{black}{When Large Language Models Meet Evolutionary Algorithms: Potential Enhancements and Challenges}}


\author[1]{\fnm{Chao} \sur{Wang}}\email{xiaofengxd@126.com}
\author[1]{\fnm{Jiaxuan} \sur{Zhao}}\email{jiaxuanzhao@stu.xidian.edu.cn}
\author*[1]{\fnm{Licheng} \sur{Jiao}}\email{lchjiao@mail.xidian.edu.cn}
\author[1]{\fnm{Lingling} \sur{Li}}\email{llli@xidian.edu.cn}
\author[1]{\fnm{Fang} \sur{Liu}}\email{f63liu@163.com}
\author[1]{\fnm{Shuyuan} \sur{Yang}}\email{syyang@xidian.edu.cn}


\affil*[1]{\orgdiv{School of Artificial Intelligence}, \orgname{Xidian University}, \orgaddress{\street{No. 2 South Taibai Road}, \city{Xi'an}, \postcode{710071}, \state{Shaanxi}, \country{China}}}


\abstract{Pre-trained large language models (LLMs) exhibit powerful capabilities for generating natural text. Evolutionary algorithms (EAs) can discover diverse solutions to complex real-world problems. Motivated by the common collective and directionality of text generation and evolution, this paper first illustrates \textcolor{black}{the conceptual parallels between LLMs and EAs at a micro level}, which includes multiple one-to-one key characteristics: \textcolor{black}{token representation and individual representation}, position encoding and fitness shaping, position embedding and selection, Transformers block and reproduction, and model training and parameter adaptation. \textcolor{black}{These parallels highlight potential opportunities for technical advancements in both LLMs and EAs.} \textcolor{black}{Subsequently, we analyze existing interdisciplinary research from a macro perspective to uncover critical challenges, with a particular focus on evolutionary fine-tuning and LLM-enhanced EAs.} \textcolor{black}{These analyses not only provide insights into the evolutionary mechanisms behind LLMs but also offer potential directions for enhancing the capabilities of artificial agents.}}



\maketitle


\begin{figure}[t]%
\centering
\includegraphics[width=\textwidth]{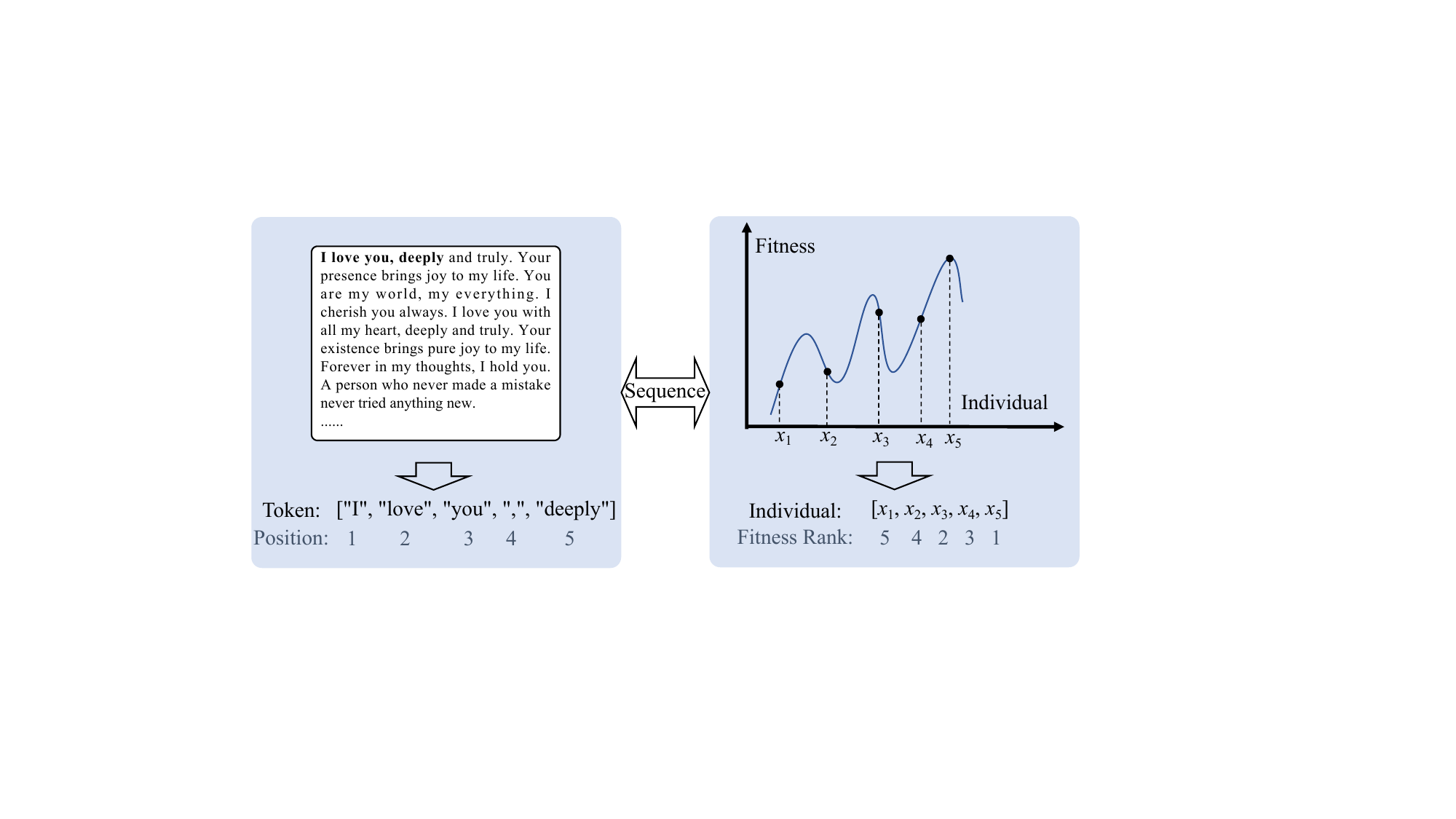}
\caption{Both tokens in a text and individuals in a population can be regarded as sequences.}\label{fig1}
\end{figure}

\textcolor{black}{Natural language processing (NLP) focuses on enabling computers to understand, generate, and process human language, covering tasks such as text generation \cite{doi:10.1126/science.aaa8685}, text segmentation \cite{9051834}, named entity recognition \cite{9174763}, sequence labeling \cite{9563226}, and relation extraction \cite{10143375}.} \textcolor{black}{Large language models (LLMs), such as generative pre-trained Transformer (GPT) \cite{radford2018improving} and bidirectional encoder representations from transformer (BERT) \cite{devlin2018bert}, primarily learn statistical patterns with temporal relations from text sequences, often in an unsupervised manner, to establish probability distributions of texts.} \textcolor{black}{These models have become foundational tools for the aforementioned NLP tasks.} \textcolor{black}{By analyzing input tokens and iteratively generating the most likely subsequent tokens, LLMs like GPT and BERT can produce coherent and contextually relevant texts. This sequence-to-sequence model with powerful understanding and generation capabilities has been employed to assist users on a variety of innovative tasks, including writing, mathematical discovery, and chemical research \cite{brown2020language,romera2023mathematical,shanahan2023role,schramowski2022large,liu2023multi,boiko2023autonomous}.} Meanwhile, training LLMs on vast text demands significant computing resources, as exemplified by ChatGPT's pre-training consumption of several thousand petaflop/s-days \cite{brown2020language}. Fine-tuning techniques have been proposed to alleviate the computational challenges of training from scratch \cite{zheng2023learn}. In a model-as-a-service scenario \cite{sun2022black}, LLMs are only accessed as inference APIs. 
\textcolor{black}{Thanks to their gradient-free nature, evolutionary algorithms (EAs) are employed to fine-tune LLMs in black-box scenarios, where they rely solely on forward propagation and do not require access to internal model gradients \cite{sun2022black}. This makes EAs a practical choice for such settings.}

\begin{figure}[htbp]%
\centering
\includegraphics[angle=90, width=0.72\textwidth]{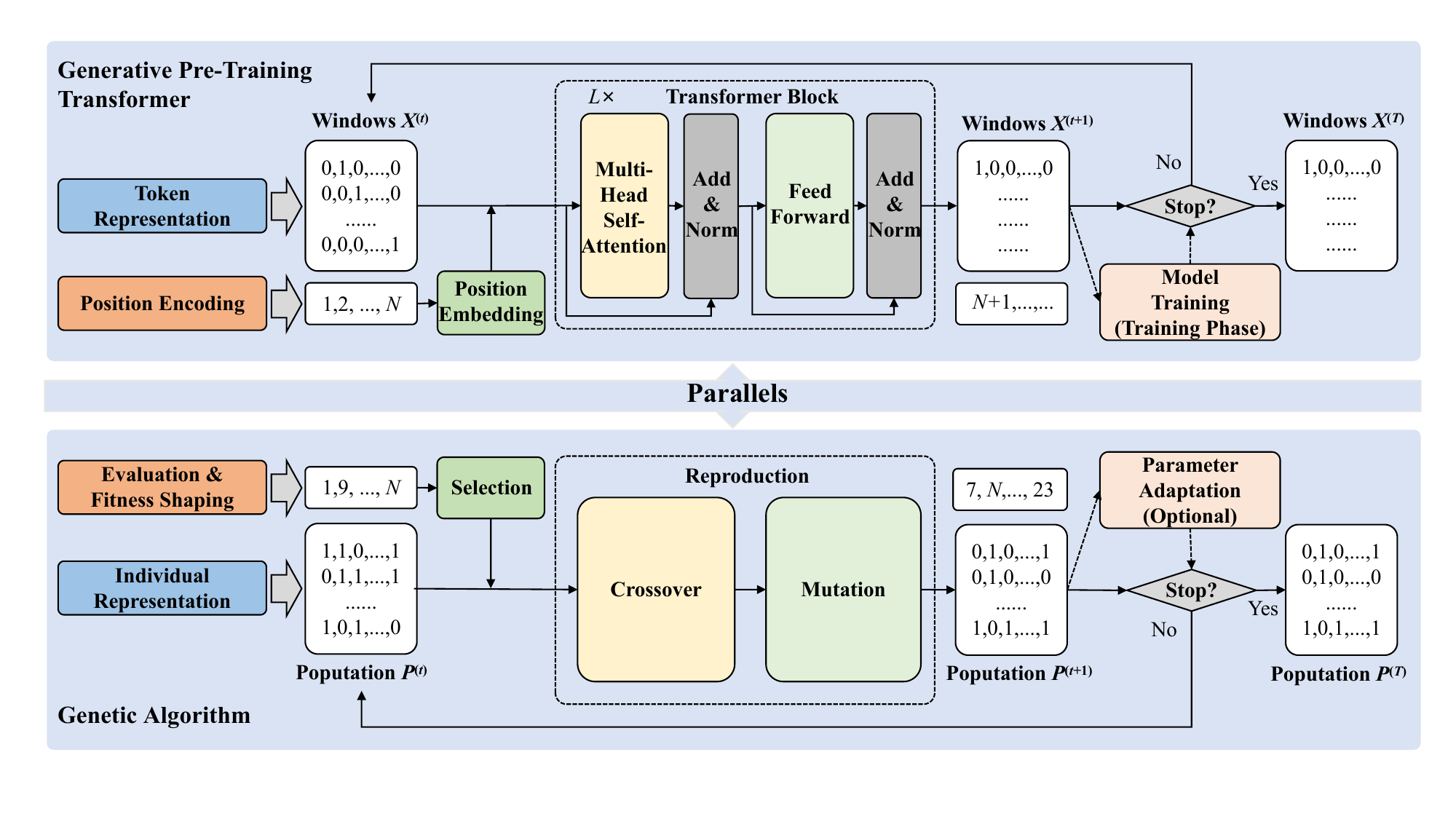}
\caption{Overview of the generative pre-training Transformer (GPT) and genetic algorithm (GA). Modules of the same color indicate \textcolor{black}{parallels}, as exemplified by the analogy between crossover in GA and attention in GPT.}\label{fig2}
\end{figure}

Drawing from biological evolution, EAs continuously maintain evolving systems (population or probability distribution) through \textcolor{black}{reproduction} and selection to explore fitness landscapes \cite{eiben2015evolutionary,kudela2022critical,doi:10.34133/research.0442}. Typical methods include the genetic algorithm (GA), evolutionary strategy (ES), and genetic programming (GP) \cite{jin2021data,miikkulainen2021biological}. In principle, only individuals and their fitness are needed to drive the evolutionary process in these approaches. Due to advancements in computational resources, EAs have provided diverse solutions to complex black-box optimization issues, such as neuroevolution \cite{stanley2019designing}, industrial design \cite{doi:10.1073/pnas.2305180120,shu2023genetic,jin2018data,li2023machine}, and natural sciences \cite{lin2023target}. Nonetheless, most EAs are task-specific, and their capabilities do not automatically increase with experience \cite{8114198,xue2023scalable,7161358,9385398}. \textcolor{black}{Recently, Transformer-enhanced EAs utilize basic Transformer models \cite{hong2023pre} to learn optimization experiences, while LLM-enhanced EAs employ well-trained LLMs \cite{brownlee2023enhancing,lehman2023evolution} to produce optimization experiences. Both approaches intend to improve the performance and generalization of EAs.}

Fig. \ref{fig1} demonstrates that text and population can be regarded as sequence data, specifically exhibiting directionality. In a text, each token corresponds to a specific position, while in a population, each individual is associated with a particular fitness rank. Text sequences have a natural directionality derived from human-defined grammatical rules. \textcolor{black}{Population evolution is also directional, primarily driven by fitness-based selection. LLMs and EAs are designed to learn or simulate such sequential data (text sequence and population sequence).} 
Fig. \ref{fig2} illustrates the process of the sequences generated by LLMs and EAs, taking the GPT \cite{radford2018improving} and GA \cite{goldberg2013genetic} as examples respectively. GPT continuously generates subsequent tokens by iterating over a context window. \textcolor{black}{The tokens in the input window are transformed by a large-scale Transformer block (multi-head self-attention and feed-forward neural network). These tokens function collectively, providing contextual information for accurate generation.} GA generates individuals with high fitness by iterating a population. Individuals in the parent population are transformed by \textcolor{black}{reproduction} operators (crossover and mutation). \textcolor{black}{These individuals exhibit collective intelligence, helping to explore the search space and seek the optimal solutions.} \textcolor{black}{During the generation process, both the context window in GPT and the population in GA are continuously updated to produce coherent texts and diverse solutions, respectively. Notably, the context window and population share a collective nature, where their constituent elements (tokens in GPT and individuals in GA) function cohesively within their respective domains.} 

Inspired by the above directionality and collective, we raise the following issues: \textcolor{black}{Are there parallels between LLMs and EAs? How can these parallels be utilized to address existing limitations and foster innovation in the coupling of LLMs and EAs? Moreover, since current interdisciplinary research focusing on either evolutionary fine-tuning or LLM-enhanced EAs remains in its infancy, what are the key challenges faced by these efforts?} To address these issues, this paper draws \textcolor{black}{conceptual} analogies between the primary characteristics of LLMs and EAs, emphasizing their common mechanisms. At the micro level, we analyze key interdisciplinary research related to each parallel, which not only supports our analogies but also offers insights for potential improvements. At the macro level, we systematically summarize evolutionary fine-tuning and LLM-enhanced EAs to reveal critical challenges. \textcolor{black}{The main contributions of this paper are as follows:}

\begin{enumerate}
    \item \textcolor{black}{At the micro level, we analyze existing research by drawing conceptual analogies between the key characteristics of LLMs and EAs, aiming to inspire novel ideas and technologies that can advance these fields.}
    \item \textcolor{black}{At the macro level, we provide the first systematic overview of evolutionary fine-tuning and LLM-enhanced EAs, highlighting key challenges for future research.}
\end{enumerate}

\textcolor{black}{The remainder of this paper is organized as follows. Section \ref{sec2} presents the conceptual parallels between LLMs and EAs from five perspectives, introducing potential technical improvements. In Sections \ref{sec3} and \ref{sec4}, we summarize evolutionary fine-tuning in black-box scenarios and LLM-enhanced EAs, respectively. Additionally, key future challenges are identified. Finally, Section \ref{sec5} summarizes the work in this paper.}


\section{\textcolor{black}{Parallels}}\label{sec2}

\begin{sidewaystable}
\caption{A comparison of large language models and evolutionary algorithms in terms of key characteristics (Chars.), \textcolor{black}{where "N/A" indicates a lack of corresponding interdisciplinary research in that area.}}\label{tab1}
\begin{tabular*}{\textwidth}{@{\extracolsep\fill}p{0.5cm}p{1.5cm}p{2.2cm}p{2.2cm}p{1.5cm}|p{1.5cm}p{2.2cm}p{2.2cm}p{1.5cm}}
\toprule%
&\multicolumn{4}{c}{Large language models}  & \multicolumn{4}{c}{Evolutionary algorithms} \\
\toprule%
Sec. & Chars. & Classic methods & Traits & \textcolor{black}{Key interdisciplinary research} & Chars. & Classic methods & Traits & \textcolor{black}{Key interdisciplinary research}\\
\midrule
\ref{sec2.1} &\textcolor{black}{Token representation} & One-hot encoding + Linear transformation & Collective, Uniqueness, finite & \textcolor{black}{\cite{sun2022black,zhao2023genetic}} &\textcolor{black}{Individual representation} & Real encoding + Random embedding \cite{LIU2024101466} & Collective, Uniqueness, Infinite and changing  & \textcolor{black}{\cite{meyerson2023language,lehman2023evolution}} \\
 \midrule
\ref{sec2.2} &Position encoding & Sine and cosine functions \cite{radford2018improving} & Uniqueness, Relativity  & \textcolor{black}{N/A} & Fitness shaping & Rank transformation, Utility function \cite{wierstra2014natural} & Uniqueness, Relativity, Directionality  & \textcolor{black}{\cite{hong2023pre}} \\
 \midrule
\ref{sec2.4} & Position embedding & Absolute, Relative, Rotary \cite{SU2024127063} & Relativity  & \textcolor{black}{N/A} & Selection & Tournament selection, Rank selection & Relativity, Directionality  & \textcolor{black}{\cite{hong2023pre}} \\
 \midrule
\ref{sec2.3} & \textcolor{black}{Transformer block} & \textcolor{black}{Multi-head self-attention + Feed-forward neural network} & \textcolor{black}{Position-insensitive, Parallelism, Sparsity, token and position information, Singleness, synergy} & \textcolor{black}{\cite{zhang2021analogous}} & \textcolor{black}{Reproduction} & \textcolor{black}{Arithmetic crossover + Uniform mutation} & \textcolor{black}{Fitness-insensitive, Parallelism, Sparsity, Individual and fitness information, Singleness, synergy}  & \textcolor{black}{\cite{hong2023pre,li2023b2opt}} \\
 \midrule
\ref{sec2.5} & Model training & \textcolor{black}{Pre-training, Fine-tuning, Reinforcement learning}  & Learning, \textcolor{black}{Exploration, Parameter space, Language space} & \textcolor{black}{\cite{sun2022black,zhao2023genetic}} & Parameter \textcolor{black}{adaptation} & \textcolor{black}{Hyper-heuristics, Pre-training, Fine-tuning, Meta-learning} & \textcolor{black}{Learning, Exploration}, \textcolor{black}{Parameter space, Search space} & \textcolor{black}{\cite{hong2023pre,lehman2023evolution,lange2023discovering,lange2023discovering2,meyerson2023language}} \\
\bottomrule
\end{tabular*}
\end{sidewaystable}


Transformer-based LLMs have developed rapidly since the introduction of the Transformer architecture \cite{vaswani2017attention}. Taking GPT as an example, LLMs mainly contain several characteristics: \textcolor{black}{token representation}, position encoding, position embedding\footnote{`Position encoding' and `position embedding' are closely related in some literature. For the sake of analogy, we have chosen to treat them separately.}, \textcolor{black}{Transformer block, and model training.} In 1950, Turing proposed a `learning machine' similar to the principles of evolution \cite{turing2009computing}. Since then, evolution-inspired computational theories have been explored and refined. Taking GA as an example, EAs include several typical characteristics: \textcolor{black}{individual representation}, fitness shaping, selection, \textcolor{black}{reproduction, and parameter adaptation}. Table \ref{tab1} lists these characteristics, classic methods, \textcolor{black}{traits, and key interdisciplinary research integrating LLMs and EAs}. Next, each subsection focuses on elucidating the corresponding characteristics of LLMs and EAs.

\subsection{\textcolor{black}{Token representation and individual representation}} \label{sec2.1}
In LLMs, the input is represented as a token sequence $X=\{x_1,...,x_N\}\in  \mathbb{R}^{N\times |V|}$, where $N$ and $|V|$ are the sizes of context window and vocabulary $V$, respectively. Each token is encoded as a high-dimensional sparse one-hot vector. Subsequently, token embeddings map token encoding sequences into a low-dimensional dense vector space \cite{radford2018improving}. For example, linear transformation applies a word embedding matrix $W_e\in \mathbb{R}^{|V|\times d_t}$ to the token encoding sequence $X$ to generate a new representation $X=XW_e$.


In EAs, the population is represented as an individual sequence $P=\{p_i,...,p_N\}\in \mathbb{R}^{N\times d}$, where $N$ and $d$ are the population size and coding dimension, respectively. \textcolor{black}{Each individual (or solution) is encoded into a data structure manipulable by genetic operators.} In numerical optimization, real encoding maps the individual into a real-valued vector.
The high dimensionality of encoding increases optimization difficulty. Many strategies are proposed to deal with the curse of dimensionality \cite{LIU2024101466,qian2017solving,tian2021evolutionary}. For example, random embedding applies a random projection matrix $W_r\in \mathbb{R}^{|V|\times d_r}$ to the population $P$ to generate a low-dimensional representation $P=PW_r$.

\textcolor{black}{Token representation can be regarded as an individual representation, which satisfies collective and uniqueness. The tokens in the context window are individuals in the population. Both token encoding and individual encoding guarantee one-to-one mapping.} 
\textcolor{black}{This analogy conceptually provides bidirectional support for interdisciplinary research. EAs using token representations can operate directly within embedded or original token spaces to find high-quality input prompts \cite{sun2022black,zhao2023genetic}.} Natural language on a finite vocabulary has demonstrated powerful representation capabilities, which may bring new opportunities for individual representation. In evolution, the decision space may be infinite, changing, and difficult to describe mathematically. \textcolor{black}{Fortunately, these complex search spaces are represented directly with the help of natural language. This flexibility enables LLM-enhanced EAs to tackle tasks that are not easily reducible to simple mathematical formulas or notations, such as paths and coding \cite{meyerson2023language,lehman2023evolution}.}


\subsection{Position encoding and fitness shaping} \label{sec2.2}
Position encoding models the dependence of tokens at different positions in the sequence. In GPT \cite{radford2018improving}, sine and cosine functions of different frequencies are adopted to encode dependencies. Each token has a unique position encoding consisting of cosine functions with different frequencies. The combination of several cosine waves contains relative distance information between tokens. However, due to the symmetry of distances, the position encoding cannot distinguish sequence direction. \textcolor{black}{Notably, Lyu et al. \cite{9671053} investigate whether models can learn directionality, emphasizing its importance for interpretability and performance.}

Fitness shaping transforms the fitness of individuals in a population to cope with selection pressure. For example, rank-based fitness shaping is commonly used in ES \cite{hansen2016cma,wierstra2014natural}. Individuals are sorted in descending order of fitness. The corresponding fitness is transformed into a set of utility values $u_1\geq u_2\geq...\geq u_N$ by a utility function. This utility function ensures invariance under fitness order-preserving transformations, which preserves the relativity and directionality of fitness. 

Both position encoding and fitness shaping share the characteristics of coding uniqueness. \textcolor{black}{The position encoding used in GPT effectively models the relative positions between tokens, although it does not explicitly capture the directionality of the sequence.} Fitness shaping describes the relative and directional ordering of individual fitness.
\textcolor{black}{Inspired by fitness shaping, the integration of sequence directionality into position encoding emerges as a noteworthy research direction. \textcolor{black}{Drawing from fitness shaping techniques in CMA-ES/NES~\cite{hansen2016cma,wierstra2014natural}, future work could design position encodings that preserve order invariance, capturing both the relative positions and directionality of tokens in text. Such methods could improve performance in tasks like long-text generation and question answering by ensuring that the model respects the temporal and contextual order of information.} Conversely, in existing Transformer-enhanced EAs \cite{hong2023pre}, fitness values are directly utilized for position encoding within the Transformer model. According to our conceptual analogy, the introduction of fitness shaping has the potential to aid Transformer-enhanced EAs in managing selective pressures.}

\subsection{\textcolor{black}{Transformer block and reproduction}} \label{sec2.3}

\textcolor{black}{A vanilla Transformer block is composed of a multi-head self-attention attention, a feed-forward neural network (FFN), residual connections, and layer normalization.} The self-attention attention mechanism performs feature transformation on the token embedding. Token embedding $X\in  \mathbb{R}^{N\times d_t}$ is transformed into query $Q=XW^Q$, key $K=XW^K$ and value $V=XW^V$, through linear transformations $W^Q\in \mathbb{R}^{d_t\times d_q}$, $W^K\in \mathbb{R}^{d_t\times d_k}$, and $W^V\in \mathbb{R}^{d_t\times d_v}$. Then, query $Q$ and key $K$ are used to calculate the attention matrix $A$ describing token relationships. Applying the $Softmax$ function to $A$ and multiplying by value $V$ yields the transformed output $X^{'}$:
\begin{equation}
A^{'}=QK^T, A=Softmax\left( \frac{A^{'}}{\sqrt{d}} \right), X^{'}=AV.
\label{eq3}
\end{equation}
The multi-head self-attention mechanism operates by combining multiple self-attention mechanisms to focus on information from different subspaces. \textcolor{black}{Many studies show that the learned attention matrix is sparse \cite{voita-etal-2019-analyzing,correia-etal-2019-adaptively,10.5555/3524938.3525019}. To address the computational cost, various improvement mechanisms have been proposed, such as sparse attention and linear attention \cite{NIU202148}.} In Transformer blocks, the FFN enhances the expressive ability of each token $x_i^{'}$ by applying a nonlinear function $f{(x_i^{'})}$. \textcolor{black}{Residual connections help alleviate the vanishing gradient problem, enabling deeper feature learning. Layer normalization stabilizes the training process and improves convergence speed.}

\textcolor{black}{Typical reproduction involves crossover and mutation.} Crossover acts on the parent population to generate new individuals. Classic real crossover operators include arithmetic crossover, simulated binary crossover\cite{deb1995simulated}, and more. We illustrate crossover's workflow using arithmetic crossover \cite{simon2013evolutionary} as an example. Any two individuals $p_i$ and $p_j$ are randomly selected from the parent population $P\in \mathbb{R}^{N\times d}$. New individual $p_i^{'}$ is a linear combination of parental genes:
\begin{equation}
p_i^{'}=a_ip_i+a_jp_j.
\label{eq4}
\end{equation}
We reformulate this process as:
\begin{equation}
p_i^{'}=0p_1+...+a_ip_i+...+a_jp_j+0p_N=[0,...,a_i,...,a_j,...]P=A_iP.
\label{eq5}
\end{equation}
Without loss of generality, $N$ individuals can be produced in a batch:
\begin{equation}
P^{'}=[A_1;A_2;...;A_N]P=AP,
\label{eq6}
\end{equation}
where $A$ is a sparse matrix determined by the selection, termed the selection matrix in the paper. \textcolor{black}{In the reproduction}, mutation applies a nonlinear perturbation $PM{(p_i^{'})}$ to each individual $p_i^{'}$ to promote  individual diversity.

Comparing (\ref{eq3}) with (\ref{eq6}), attention and crossover share a similar mathematical representation, exhibiting parallelism and sparsity. \textcolor{black}{Attention does not explicitly model token positions. Similarly, crossover inherently does not consider individual fitness.} The attention and selection matrices play analogous roles: one determines token feature combinations, while the other governs parent genetic combinations. The attention matrix is parameterized based on token embeddings, while the selection matrix is heuristically built on individual relationships. \textcolor{black}{Additionally, (\ref{eq3}) demonstrates that the input and output of attention occupy distinct latent spaces. However, (\ref{eq6}) reveals the population keeps the same representation space across crossover.} \textcolor{black}{This analogy highlights opportunities to borrow ideas from attention mechanisms to improve crossover operators and vice versa, as demonstrated in recent studies~\cite{li2023b2opt,zhang2021analogous}.}

Both the FFN and mutation operate independently on each singleton (token or individual) and can be processed in parallel. Existing works \cite{dong2021attention,zhang2021analogous} show that the removal of FFNs degrades Transformer performance, emphasizing the crucial role of both attention and FFNs. Analogously, the synergistic effect of crossover and mutation results in the super efficiency of EAs \cite{zhou2019evolutionary,hassanat2019choosing}. \textcolor{black}{While FFNs and mutation are not mathematically equivalent, their functional roles in maintaining performance are conceptually similar. This analogy suggests potential opportunities to enhance FFNs by introducing controlled randomness, inspired by the role of mutation in generating diversity within populations in EAs.}

\textcolor{black}{Zhang et al. \cite{zhang2021analogous} first analogized Transformer blocks to reproduction, utilizing dynamic local populations in EAs to enhance Vision Transformers. Similarly, Li et al. \cite{li2023b2opt} modeled crossover and mutation using attention and FFN mechanisms, respectively, to develop Transformer-enhanced EAs. These efforts highlight the functional parallels between Transformer components and EA operations, confirming our conceptual analogy and demonstrating the potential for progress through idea sharing between advanced Transformers and reproduction.}

\subsection{Position embedding and selection} \label{sec2.4}

Position embeddings integrate positional information into the attention mechanism, using absolute, relative, and rotary techniques, to capture sequential dependencies and contextual relationships within token sequences \cite{SU2024127063}. A typical absolute position embedding is the sinusoidal position embedding, adding \textcolor{black}{position information encoded by sine and cosine functions} to token embeddings. For any two tokens $x_t$ and $x_s$, with position information $p_t$ and $p_s$, the attention matrix is expressed as:
\begin{align}
A_{t,s}&=Q_t^TK_s=(x_t+p_t)^TW_Q^TW_K(x_s+p_s) \\
&=x_t^TW_Q^TW_Kx_s+x_t^TW_Q^TW_Kp_s+p_t^TW_Q^TW_Kx_s+p_t^TW_Q^TW_Kp_s.
\label{eq7}    
\end{align}
Due to $W_Q^TW_K$, the token relative information is destroyed \cite{yan2019tener}. Several models like T5 \cite{raffel2020exploring}, Transformer-XL \cite{dai2019transformer}, TENER \cite{yan2019tener}, and DeBERTa \cite{he2020deberta} have integrated relative positional information into the attention matrix to address this limitation. For example, T5 directly adds token offsets to attention weights:
\begin{equation}
A_{t,s}=Q_t^TK_s+r_{b(t-s)}=x_t^TW_Q^TW_Kx_s+r_{b(t-s)}.
\label{eq8}       
\end{equation}
Rotary position embedding incorporates relative position information through token embedding rotation \cite{SU2024127063}:
\begin{equation}
A_{t,s}=x_tW_QR_{t-s}W^T_Kx_s^T=x_tW_QR_{t}(R_{s}^TW^T_Kx_s^T),
\label{eq9}       
\end{equation}
which achieves a unification of absolute and relative position embeddings.

Based on fitness information, selection operators such as tournament and rank selection are designed to identify excellent parents for crossover \cite{simon2013evolutionary}. These selection methods can be viewed as heuristics for building the selection matrix. In binary tournament selection, for instance, the selection matrix $A$ is randomly created based on fitness comparisons. In basic differential evolution \cite{das2010differential}, multiple individuals are randomly chosen for differential operations, which influences the composition of the selection matrix $A$. In OpenAI-ES \cite{salimans2017evolution}, each individual is sampled from a multivariate Gaussian distribution with mean $\frac{1}{N}\sum_1^Nu_ip_i$ and covariance $\sigma^2I$:
\begin{equation}
p_i^{'}=\frac{1}{N}\sum_1^Nu_ip_i+N(0,\sigma^2I),
\label{eq10}       
\end{equation}
where $u_i$ is the utility function of $i$-th parent individual $p_i$. The first term is crossover and the second is mutation. The crossover is rewritten as:
\begin{equation}
\frac{1}{N}\sum_1^Nu_ip_i=\left[\frac{1}{N}u_1,...,\frac{1}{N}u_N\right]P=A_iP,
\label{eq11}       
\end{equation}
where each row of the selection matrix $A_i$ is determined by the parent's utility value. In OpenAI-ES, the selection matrix has identical rows due to the same genetic material from parents assigned to each individual. Furthermore, in NSGAII \cite{deb2002fast}, the selection matrix A is constructed using non-dominated sorting and crowding distance, considering individual relationships in the objective space.

By operating on matrix $A$, position embedding and selection add position and fitness information to the attention mechanism and crossover, respectively. The selection notably steers the population towards enhanced fitness. \textcolor{black}{Individuals with higher fitness are preferentially selected for crossover, which considers the directionality of individual fitness, fostering a more adaptive population. }
\textcolor{black}{Standard position embedding effectively captures the relative positions between tokens but does not explicitly encode the sequential order of tokens. Current efforts introduced task-specific supervision during training to assist LLMs in comprehending the sequential relationships among tokens. For example, GPT \cite{radford2018improving} employs a masked multi-head attention mechanism, ensuring that the output at each position is solely determined by preceding tokens. This approach guarantees a unidirectional information flow, forcing the self-attention mechanism to consider only the past context. Masked LLMs like BERT \cite{wagner2020position} learn contextual representations by predicting masked tokens, necessitating a focus on the entire textual context in both directions rather than just a unilateral one. Inspired by the directionality of fitness considered in selection, introducing token order directly into position embedding may enhance the generative capabilities of LLMs.}


The attention matrix is influenced by both token and position relations. The selection matrix is usually designed based on fitness relations. In complex fitness landscapes, additional considerations such as genetic similarity among individuals are factored into the selection matrix. For instance, in multi-modal optimization with multiple global optima \cite{10.1145/1143997.1144200,das2011real}, individual distances within the search space are employed as a criterion to preserve population diversity during selection. \textcolor{black}{This ensures that the population does not converge prematurely to a single optimum. The selection, in this context, incorporates both individual relations (e.g., distances between individuals) and fitness relations (e.g., fitness rank).}
\textcolor{black}{Essentially, the similarity between attention and selection matrices in handling relational information stems from the analogy drawn to tokens and individuals, as well as to positions and fitness. Existing Transformer-enhanced EA merges individual embeddings with fitness embeddings, echoing how token embeddings are combined with positional embeddings \cite{hong2023pre}. This practice serves as a support for our conceptual analogy. Advanced attention mechanisms, such as those incorporating rotational positional embedding \cite{SU2024127063}, have the potential to enhance the performance of this operation in Transformer-enhanced EAs.}


\subsection{Model training and parameter \textcolor{black}{adaptation}} \label{sec2.5}

\textcolor{black}{Model training typically begins with unsupervised pre-training, modeling natural language on a vast amount of text. This is followed by fine-tuning, which adjusts the model to downstream tasks.
} Given a token sequence $X=\{x_1, ..., x_N\}$, a unidirectional LLM estimates a conditional probability distribution $P(x_i|x_1,...,x_{i-1};\theta)$ to generate subsequent tokens. For example, In GPT's pre-training \cite{radford2018improving}, the goal of language modeling is to maximize the log-likelihood:
\begin{equation}
 L^{PT} = \sum_{i=1}^N \log P(x_i|x_1,...,x_N;\theta),
\label{eq12}       
\end{equation}
where $N$ is the context window size and $\theta$ is the model parameter. \textcolor{black}{During fine-tuning, the optimization objective is a weighted sum of pre-training loss $L^{PT}$ and fine-tuning loss $L^{FT}$ \cite{radford2018improving}:}
\begin{equation}
L=L^{FT}+\mu L^{PT}.
\label{eq14}       
\end{equation}
Hyperparameter $\mu\in [0,1]$ determines the trade-off across losses. \textcolor{black}{Furthermore, reinforcement learning \cite{du2023guiding} fine-tunes LLMs by optimizing outputs' overall performance (rewards) to generate high-quality responses continuously. Evolutionary fine-tuning is proposed for black-box cases with inaccessible gradients and limited resources. These methods typically guide LLMs to generate the desired output by automatically constructing prompts directly within the input sequence \cite{sun2022black,zhao2023genetic}.}

In GA, given parent population $P=\{p_1,...,p_N\}$, offspring are sampled from an implicit conditional probability distribution $P(p_i|p_1,...,p_N)$, which is induced by genetic operators including selection, crossover, and mutation \cite{meyerson2023language}. \textcolor{black}{Genetic operator parameters are often determined through repeated experiments or adaptively updated using hyper-heuristic strategies \cite{burke2013hyper}.} In ES, the probability distribution $P(p|p_1,...p_N;\theta)$ over the parent population is employed to produce offspring. For example, in CMAES \cite{hansen2016cma}, the parameters (mean and covariance matrix) of a multivariate Gaussian distribution are adapted by maximizing the log-likelihood:
\begin{equation}
 \sum_{i=1}^N u_i\log P(p_{i:N};m);
\sum_{i=1}^N u_i\log P\left(\frac{p_{i:N}-m}{\sigma};C\right),
\label{eq13}       
\end{equation}
where \textcolor{black}{$p_{i:N}$ refers to the $i$th ranked individual in a population with $N$ individuals based on fitness. The first term is the mean update, while the second term is the ranking-$N$ update of the covariance matrix. Recently, Transformer-enhanced EAs adaptively updated parameters from optimization experiences on a set of optimization tasks, improving the generalization ability of EAs on new tasks. Common methods include pre-training \cite{hong2023pre,lehman2023evolution} and meta-learning \cite{lange2023discovering,lange2023discovering2}. In pre-training, optimization experiences for multi-objective optimization \cite{hong2023pre} consist of the population and their fitness generated by multi-objective EAs on numerous benchmarks. Optimization experiences for GP \cite{lehman2023evolution} include incremental changes in files submitted by humans to version control systems like Github. Meta-learning  \cite{lange2023discovering,lange2023discovering2} adaptively updates parameters by optimizing the average performance of EAs across a set of tasks. Regrettably, no comprehensive study has compared these two methods within a single framework. \textcolor{black}{Furthermore, well-trained LLMs are directly utilized as reproduction operators with human-like experience \cite{lehman2023evolution,ding2024quality}.} Prompts based on historical populations are constructed to guide LLMs in generating the desired output population \cite{meyerson2023language,lehman2023evolution}.}


Despite differing implementation strategies, LLMs and EAs converge on a shared fundamental objective: revealing the underlying probability distributions within data, thereby facilitating the learning and exploration of knowledge. \textcolor{black}{In pre-training and supervised fine-tuning, LLMs construct conditional probability distributions through the accurate prediction of tokens, learning vast pre-existing knowledge. Reinforcement learning adjusts the LLM parameters based on the rewards. \textcolor{black}{Evolutionary fine-tuning automatically searches for high-quality prompts or configurations to improve output quality \cite{sun2022black,klein2023structural}.} These two paradigms explore new knowledge specific to the target task. Learning and exploration jointly ensure the generative and generalization abilities of LLMs.} \textcolor{black}{EAs shape probability distributions based on fitness evaluated via real-time individual-environment interaction. In Transformer/LLM-enhanced EAs, models learn from existing optimization experiences or human-like experiences, endowing EAs with powerful learning capabilities.}
Additionally, LLM training works in the parameter space, while evolutionary fine-tuning extends to the language space. EAs operate both in the search space (e.g., GA) and the parameter space (e.g., CMAES). The aforementioned analogy provides a reasonable motivation for interdisciplinary research: merging the exceptional learning capabilities of LLMs with the remarkable exploration abilities of EAs can foster advancements in their respective fields. 


\textcolor{black}{In (\ref{eq14}), the hyperparameter $\mu$ is carefully tuned manually, as it affects the generalization ability of LLMs. The loss trade-off is modeled as a multi-objective optimization problem \cite{sener2018multi,lin2019pareto}. Applying advanced multi-objective EAs aids in creating stronger supervised fine-tuning paradigms. Recent studies \cite{salimans2017evolution} show that ES has advantages over gradient-based reinforcement learning for long episodes with very many time steps. ES is promising as an alternative to reinforcement learning for training LLMs in multi-turn dialogue systems.}

\textcolor{black}{The generalization of the Transformer-enhanced EAs is influenced by optimization experience, which involves a set of historical optimization tasks. The similarity between historical and new tasks determines the effectiveness of optimization experience utilization, echoing the motivation behind evolutionary transfer optimization (ETO) \cite{8114198}. Benchmarks for ETO \cite{xue2023scalable} can potentially serve as optimization experiences for training in diverse transfer scenarios. Additionally, using algorithms, human expertise, or LLMs to generate optimization experiences across various benchmarks is critical for expanding the application scope.
}

\subsection{\textcolor{black}{Summary}}

\begin{figure}[htbp]%
\centering
\includegraphics[angle=90, width=0.67\textwidth]{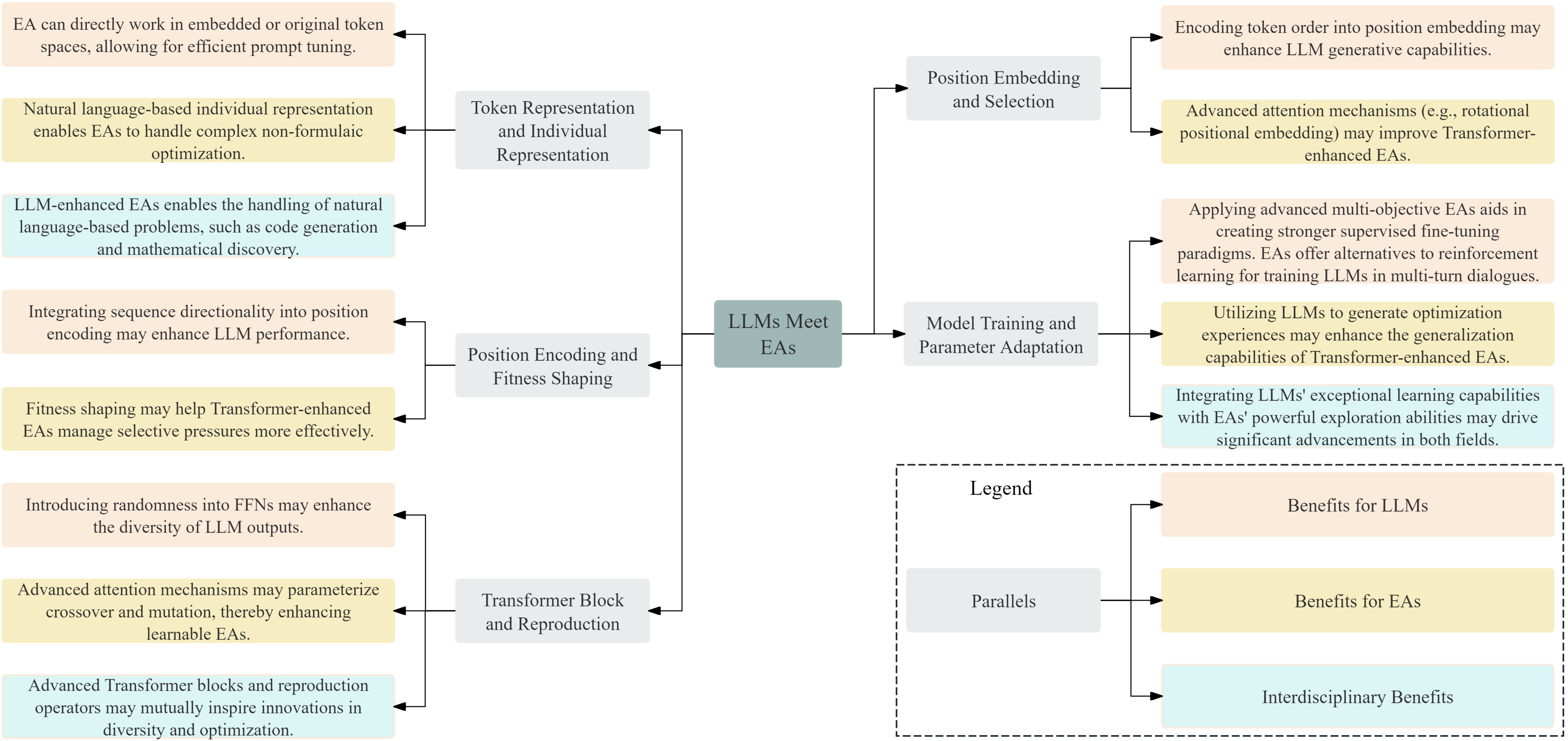}
\caption{\textcolor{black}{Conceptual parallels between large language models and evolutionary algorithms: inspiring novel ideas and technical advancements.}}\label{parallels}
\end{figure}

\textcolor{black}{Despite their independent development, LLMs and EAs share certain conceptual parallels. The parallels have inspired novel ideas and technical advancements, as outlined in Fig. \ref{parallels}.} \textcolor{black}{From a macro perspective, the parallels between LLMs and EAs provide a conceptual framework that can inspire the development of artificial agents capable of learning from established knowledge while continuously exploring new knowledge. For instance, recent studies have demonstrated that integrating EAs with LLMs can significantly enhance their performance and expand their application scope \cite{sun2022black,zhao2023genetic,meyerson2023language,li2023b2opt,hong2023pre,lehman2023evolution,lange2023discovering,lange2023discovering2,zhang2021analogous}.} However, a unified paradigm with one-to-one key feature correspondence has not emerged. \textcolor{black}{We stress that the analogy aims not to validate these parallels mathematically but to provide researchers with a pathway to enhance current technical studies.} In existing efforts to integrate EAs and LLMs, evolutionary fine-tuning in black-box scenarios and LLM-enhanced EAs are receiving increasing attention. Next, this paper provides a comprehensive review of them to identify key challenges.

\section{Evolutionary fine-tuning in black-box scenarios} \label{sec3}

The fine-tuning reduces the risk of data leakage and avoids the huge computational cost of training a model from scratch \cite{zheng2023learn}. EA is widely used to fine-tune LLMs in complex scenarios due to their flexibility. Evolutionary model tuning adjusts the model's weights or architecture \cite{choong2023jack,klein2023structural,anonymous2023knowledge}, requiring a deep understanding of LLM internals. However, real-world constraints like limited computing resources and access restrictions can hinder this process. \textcolor{black}{ In contrast, evolutionary prompt tuning \cite{sun2022black} and evolutionary self-tuning \cite{singh2022explaining,fernando2023promptbreeder,li2023spell,chen2023evoprompting,zhang2023auto} primarily focus on modifying the model's input to enhance performance on specific tasks, requiring access to no internal information. These evolutionary fine-tuning techniques in black-box scenarios are gaining attention for their low cost, as detailed in Tables \ref{tab3} and \ref{tab4}.}



\begin{sidewaystable}
\caption{A comprehensive summary of evolutionary prompt tuning, highlighting its key characteristics: decision variable and its traits, objective and its traits, adopted method, introducing new models, retraining, and internal access.}\label{tab3}
\centering
\tiny
\begin{tabular*}{\linewidth}
{p{1.5cm}|p{1.5cm}|p{1.5cm}|p{1.5cm}|p{1.5cm}|p{4cm}|p{1.5cm}|p{1.5cm}|p{1.5cm}}
\toprule%
Literature & Decision variable & Variable traits & Objective & Objective traits & Adopted methods & New model & Retraining & Internal access \\
\midrule
BBT \cite{sun2022black} & Prompt embedding & Continuous & Loss & Single-objective & Random embedding, CMAES & No & No & No \\
\midrule
BBTv2 \cite{sun2022bbtv2} & Prompt embedding & Continuous & Loss & Single-objective & Divide-and-conquer, Random embedding, CMAES & No & No & Yes \\
\midrule
Textual inversion \cite{fei2023gradient} & Prompt embedding & Continuous & Loss & Single-objective & Subspace decomposition, Random embedding, CMAES & No & No & No \\
\midrule
SNPE/ABC-SMC \cite{shen2023reliable} & Prompt embedding & Continuous & Loss & Single-objective & Variational inference, Random embedding, CMAES & No & No & No \\
\midrule
PCT \cite{qi2023prompt} & Prompt embedding & Continuous & Loss & Single-objective & Prompt-Calibrated Tuning, Whole-word mask, CMAES & No & No & Yes \\
\midrule
BBT-RGB \cite{sun2023make} & Prompt embedding & Continuous & Loss & Single-objective & Divide-and-conquer, Random embedding, COBYLA, CMAES & No & No & Yes \\
\midrule
BSL \cite{zheng2023black} & Prompt embedding & Continuous & Loss & Single-objective & Subspace learning, Random embedding, CMAES & No & No & No \\
\midrule
GDFO \cite{han2023gradient} & Prompt embedding & Continuous & Loss & Single-objective & Knowledge distillation, Random embedding, CMAES & Student model & Yes & No \\
\midrule
FedBPT \cite{sun2023fedbpt} & Prompt embedding & Continuous & Multi-client loss & Single-objective & Federated CMAES & No & No & No \\
\midrule
GAP3 \cite{zhao2023genetic} & Prompt & Discrete & Performance score, Predicted probability & Multi-objective & Multi-level evaluation, Genetic algorithm & No & No & No \\
\midrule
GrIPS \cite{prasad2022grips} & Prompt & Discrete & Accuracy, Entropy & Multi-objective & Weighted sum, Genetic algorithm & No & No & No \\
\midrule
ClaPS \cite{zhou2023survival} & Prompt & Discrete & Loss & Single-objective & Clustering and pruning, Evolutionary algorithm & No & No & No \\
\midrule
Attacks \cite{lapid2023open} & Prompt & Discrete & Cosine similarity & Single-objective & Fitness approximation, Genetic algorithm & No & No & No \\
\midrule
BPT-VLM \cite{yu2023black} & Text-image prompt embedding & Continuous & Loss & Single-objective & Random embedding, MMES, MAES, CMAES & No & No & No \\
\botrule
\end{tabular*}
\end{sidewaystable}

\begin{sidewaystable}
\caption{A comprehensive summary of evolutionary self-tuning, highlighting its key characteristics: decision variable and its traits, objective and its traits, adopted method, introducing new models, retraining, and internal access.}\label{tab4}
\centering
\begin{tabular*}{\linewidth}{p{1.9cm}|p{1.5cm}|p{1.5cm}|p{1.6cm}|p{1.5cm}|p{4cm}|p{1.2cm}|p{1.3cm}|p{1.2cm}}
\toprule%
Literature & Decision variable & Variable traits & Objective & Objective traits & Adopted methods & New model & Retraining & Internal access \\
\midrule
iPrompt \cite{singh2022explaining} & Prompt & Discrete & Render function & Single-objective & LLM-based genetic operators, Rank-based selection, Exploration & No & No & No \\
\midrule
Promptbreeder \cite{fernando2023promptbreeder} & Task-mutation prompt & Discrete & Performance score & Single-objective & LLM-based genetic operators, Genetic algorithm & No & No & No \\
\midrule
Auto-Instruct \cite{zhang2023auto} & Instruction & Discrete & Predicted score & Single-objective & LLM-based genetic operators, Rank-based selection model & Selection model & Yes & No \\
\midrule
SPELL \cite{li2023spell} & Prompt & Discrete & Classification accuracy & Single-objective & LLM-based genetic operators, Genetic algorithm & No & No & No \\
\midrule
EVOPROMPT \cite{chen2023evoprompting}& Prompt & Discrete & Performance score & Single-objective & LLM-based genetic operators, Genetic algorithm, Differential evolution & No & No & No  \\

\botrule
\end{tabular*}
\end{sidewaystable}

\begin{figure}[t]%
\centering
\includegraphics[width=\textwidth]{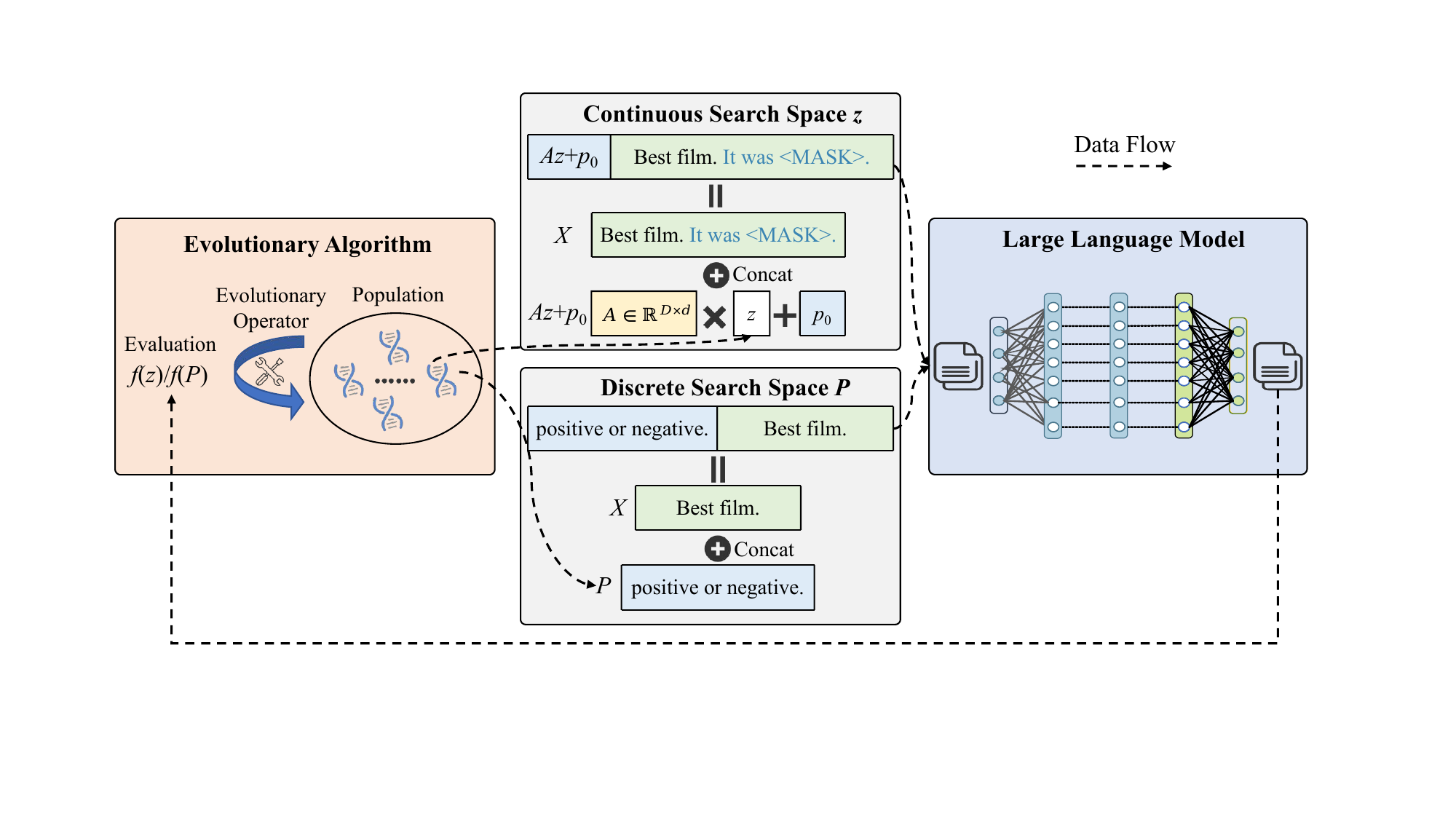}
\caption{Basic workflow of evolutionary prompt tuning. Evolutionary algorithms are utilized to efficiently search for optimal discrete prompts or continuous prompt embeddings thereby boosting the performance of large language models on downstream tasks.}\label{fig3}
\end{figure}

As shown in Fig. \ref{fig3}, evolutionary prompt tuning enhances model generation quality in few-shot or zero-shot settings by searching input prompts. EAs are employed to find prompts to maximize task performance \cite{sun2022black}, relying solely on LLM inference results. Current approaches are categorized as continuous and discrete prompt tuning. Continuous prompt tuning uses continuous EAs like CMAES to refine prompt embeddings. To enrich the information within the embedding space, various decomposition strategies are incorporated such as divide-and-conquer, subspace learning, and others \cite{sun2022bbtv2,fei2023gradient,qi2023prompt,zheng2023black}. Meanwhile, techniques like knowledge distillation, variational inference, and federated learning are used to boost search efficiency, improve generalization, and enhance security \cite{shen2023reliable,sun2023make,han2023gradient,sun2023fedbpt}. Continuous prompts require embedding space access, unsuitable for strict black-box settings. Discrete prompt tuning directly searches the prompt space using discrete EAs, in which custom genetic operators heuristically modify prompts \cite{zhao2023genetic,prasad2022grips}. Zhou et al. \cite{zhou2023survival} clustered and pruned the discrete search space to target promising prompt regions, addressing combinatorial explosion. In addition, evolutionary prompt tuning is also used in adversarial attacks and multi-modal learning \cite{lapid2023open,yu2023black}, generating effective attacks and diverse prompts. Recently, LLMs, with strong generative capacity, act as genetic operators in EAs, creating high-quality prompts \cite{singh2022explaining,fernando2023promptbreeder,li2023spell,chen2023evoprompting}, termed self-tuning in this paper. In addition to prompt generation, LLMs can serve as versatile prompt selectors for out-of-domain tasks \cite{zhang2023auto}. Self-tuning works in a flexible language space, independent of parameter updates.

Evolutionary model tuning targets parameter space, while evolutionary prompt tuning and self-tuning focus on language (search) space. Compared to model training involving multiple gradient descents, evolutionary black-box tuning is highly cost-effective. Current research focuses on model evolution within the language space. \textcolor{black}{In open environments, complex tasks may require self-coevolution in language and parameter spaces, posing challenges like resource management, catastrophic forgetting, fitness assessment, and security issues. Efficient resource management strategies help save costs in continuous evolution. Finding a balance between new and old knowledge mitigates catastrophic forgetting. Designing collaborative evaluation strategies for language and parameter spaces tailored to specific tasks is essential. \textcolor{black}{As self-evolving systems continue to advance, the development of robust security assessment mechanisms may become critical to address potential ethical challenges.}} In addition, integrating text, images, audio, and video is increasingly crucial \cite{fei2022towards}. Evolutionary multimodal fusion techniques \cite{yu2023black,anonymous2023knowledge} offer a promising path to unifying diverse information, thereby expanding the applicability and versatility of evolutionary fine-tuning.


\section{Large language model-enhanced evolutionary algorithm} \label{sec4}

\begin{figure}[t]%
\centering
\includegraphics[width=\textwidth]{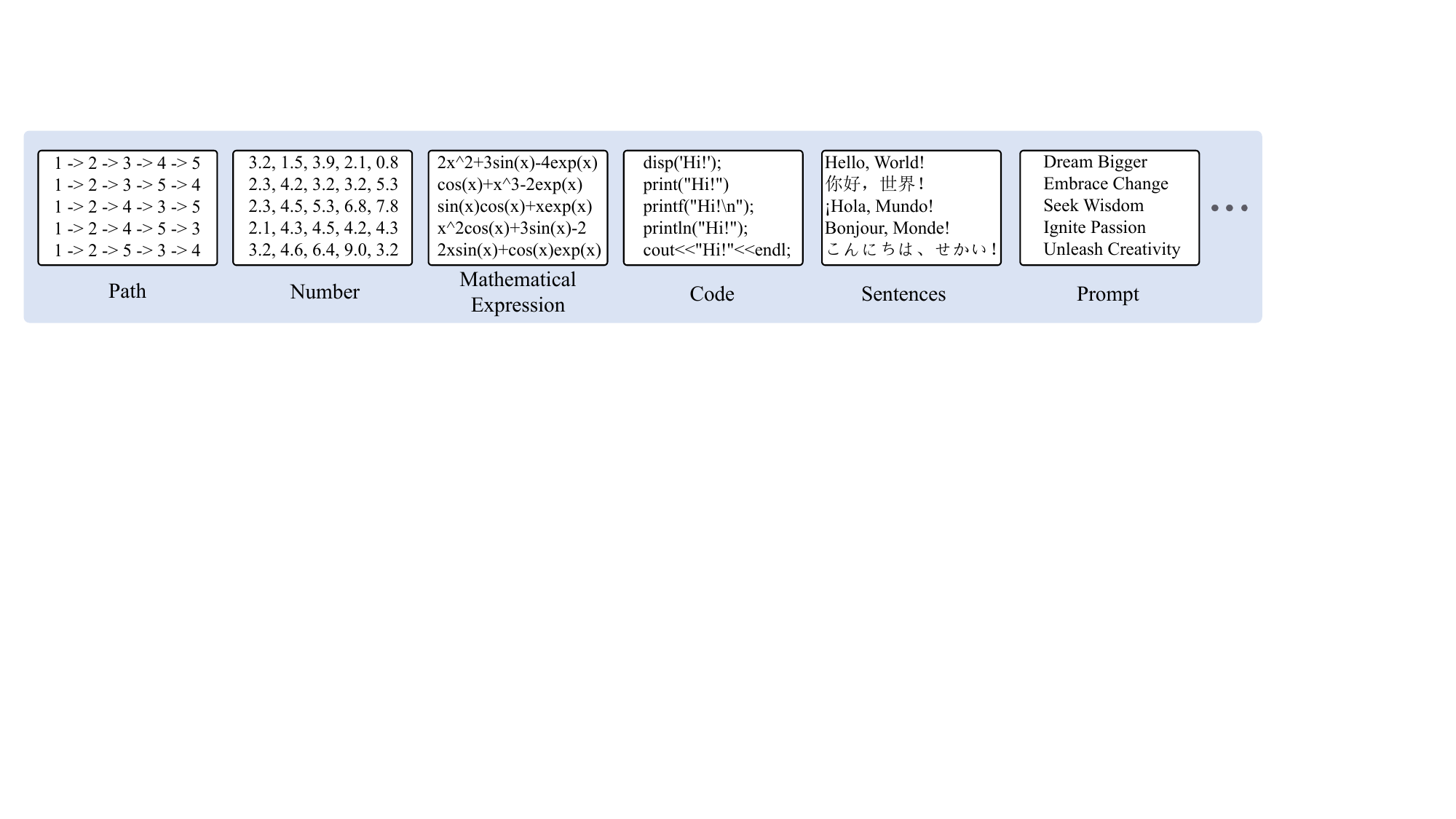}
\caption{Various complex \textcolor{black}{individual representations} can be represented directly using natural language descriptions, such as paths, numbers, mathematical expressions, code, sentences, and prompts.}\label{fig4}
\end{figure}

Fig. \ref{fig4} illustrates how complex \textcolor{black}{individual representations} are represented via flexible natural language. Language-represented populations can be directly processed by LLMs with strong text comprehension and generation skills. Table \ref{tab5} summarizes the LLM-enhanced EAs, where LLMs are employed as the reproduction and mutation operator. These methods maintain the population via LLM-based evolutionary operators to find diverse solutions to complex real-world challenges.

\begin{table}[t]
\caption{Large language models-enhanced evolutionary algorithms. Large language models are employed as the reproduction and mutation operators in evolutionary algorithms.}\label{tab5}
\begin{tabular*}{\textwidth}{@{\extracolsep\fill}p{3cm}|p{4.5cm}|p{4.5cm}}
\toprule%
& Reproduction & Mutation \\
\midrule
Prompt construction  & Problem description (optional), Population, Task instructions & Individual, Task instructions \\
\midrule
Operation space  & Language space & Language space, Parameter space \\
\midrule
Evolutionary algorithms & Hill climbing, Genetic algorithm, Genetic programming, Quality diversity, MOEA/D, Local search & Genetic algorithm, Genetic programming, Quality diversity \\
\midrule
Applications  & Function search\cite{romera2023mathematical}, Combinatorial optimization\cite{liu2023large}, Reward function optimization\cite{ma2023eureka}, Automatic machine learning\cite{nasir2023llmatic,zheng2023can,wang2023graph,zhang2023using,zhang2023automl}, Multi-objective optimization\cite{liuu2023large}, Prompt turning\cite{yang2023large}, Text generation\cite{xiao2023enhancing,bradley2023quality}, Game design\cite{lanzi2023chatgpt,sudhakaran2023mariogpt}, Materials science\cite{jablonka202314}, Image generation\cite{meyerson2023language}, Algorithm design\cite{liu2024example} & Code generation\cite{brownlee2023enhancing,lehman2023evolution}, Data management\cite{chen2023seed}, Fuzzing\cite{xia2023universal} \\
\botrule
\end{tabular*}
\end{table}

LLM-based reproduction enables the LLMs to derive offspring from parents based on prompts. Prompts usually consist of a problem description (optional), parent population, and task instructions. LLMs apply task instructions to the parent population, generating offspring \cite{meyerson2023language}. Romera-Paredes et al. \cite{romera2023mathematical} introduced a program search method for mathematical reasoning, where LLMs create multiple programs from parents. Fitness, expressed via numerical values, training logs, and human feedback, is also integrated into prompts to guide the reproduction process \cite{liu2023large,liuu2023large,ma2023eureka,yang2023large,jablonka202314,brownlee2023enhancing,lehman2023evolution,chen2023seed,xia2023universal,lanzi2023chatgpt,sudhakaran2023mariogpt,xiao2023enhancing,bradley2023quality,liu2024example}. For example, in automated learning, training logs act as fitness for finding efficient architectures and hyperparameters \cite{nasir2023llmatic,zheng2023can,wang2023graph,zhang2023using,zhang2023automl}. Reproduction using LLMs operates directly in language space, without needing extensive parameter access, resulting in cost savings.

LLMs can also be viewed as mutation operators affecting a single individual. Mutation prompts typically include individual and task instructions. For instance, in the data management strategy SEED \cite{xia2023universal}, LLMs branch an initial code fragment into multiple new code fragments based on task instructions. In addition, Lehman et al. \cite{lehman2023evolution} introduced a mutation operator \textit{diff} that acts on the parameter space. The \textit{diff} model performs parameter updates in an autoregressive manner, learning incremental changes to files. Given a parent code, the \textit{diff} can simulate the modification behavior of human programmers to generate new code. 

\textcolor{black}{Current methods primarily operate in the language space, offering high flexibility and low cost. However, when model parameters are accessible, designing efficient genetic operators in both parameter and language spaces deserves deeper investigation for potential improvements. During evolution, LLMs must address the exploration-exploitation challenge.} Exploration encourages the generation of novel and diverse outputs. While exploitation prioritizes outputs that are highly relevant to the given context, potentially sacrificing creativity. Striking a balance between these two strategies determines the ability of LLMs to autonomously acquire new knowledge. \textcolor{black}{Evolutionary multi-objective optimization \cite{10.1145/3425501} promises to provide a set of solutions with different trade-offs.}

\section{Conclusion} \label{sec5}

\textcolor{black}{LLMs and EAs have spurred innovation across interdisciplinary domains, with their synergistic integration holding the potential to realize the evolutionary learning machines envisioned by Turing \cite{turing2009computing}.} \textcolor{black}{This paper demonstrates the conceptual parallels between LLMs and EAs from five aspects, initially indicating that analogies can potentially spark a new artificial intelligence paradigm integrating LLM's learning abilities with EA's exploratory capabilities.} Recently, LLMs have shown promise in utilizing principles of evolution \cite{meyerson2023language,romera2023mathematical,lehman2023evolution,ma2023eureka,singh2022explaining,fernando2023promptbreeder,li2023spell,chen2023evoprompting,zhang2023auto}. The exponential growth in computing power enables \textcolor{black}{ large models combined with evolutionary mechanisms to perform reasoning in complex environments.} \textcolor{black}{Building on these developments, our work highlights promising directions for advancing current research and identifies critical challenges for future progress.}

\bibliography{sn-bibliography}

\end{document}